\documentclass[letterpaper, 10 pt, conference]{ieeeconf}  

\IEEEoverridecommandlockouts                              

\usepackage{multirow}
\usepackage{graphicx}
\usepackage{array}
\usepackage{booktabs}
\usepackage{stmaryrd}
\usepackage{multirow} 
\usepackage{color}
\usepackage{amsmath}
\usepackage{arydshln}
\usepackage{makecell}
\usepackage{pifont}
\usepackage{tabularx}

\usepackage[table]{xcolor} 
\usepackage{tikz}
\definecolor{lightblue}{rgb}{0.8, 0.9, 1.0} 

\usepackage{amsmath, amsfonts, amsthm, stackrel, amssymb}
\usepackage{float}
\usepackage{stfloats}
\usepackage{colortbl}
\usepackage{hyperref} 
\usepackage{xcolor}
\usepackage{hyperref}
\usepackage{cite}
\usepackage{caption}

\newcommand{\eg}{\emph{e.g., }}

\newcommand{\projname}{RANGER}
\newcommand{\projnamebold}{\textbf{\projname}}

\makeatletter
\let\AB@affilsepx=\AB@affilsepx@ieee 
\makeatother

\title{\LARGE \bf
\projname: A Monocular Zero-Shot Semantic Navigation Framework through Visual Contextual Adaptation}

\author{Ming-Ming Yu$^{1,2}$, Yi Chen$^{3}$, Börje F. Karlsson$^{2\dagger}$, and Wenjun Wu$^{1,4\dagger}$%
\thanks{$^{1}$Beihang University; $^{2}$Beijing Academy of Artificial Intelligence; $^{3}$Institute of Automation, Chinese Academy of Sciences; $^{4}$Hangzhou International Innovation Institute, Beihang University. $^{\dagger}$Corresponding authors. Emails: {\tt\small \{mingmingyu, wwj\}@buaa.edu.cn}, {\tt\small  borje@baai.ac.cn}}%
}
\begin{document}

\maketitle
\thispagestyle{empty}
\pagestyle{empty}

\begin{abstract}
Efficient target localization and autonomous navigation in complex environments are fundamental to real-world embodied applications.
While recent advances in multimodal foundation models have enabled zero-shot object goal navigation, allowing robots to search for arbitrary objects without fine-tuning, existing methods face two key limitations: (1) heavy reliance on ground-truth depth and pose information, which restricts applicability in real-world scenarios; and (2) lack of visual in-context learning (VICL) capability to extract geometric and semantic priors from environmental context,  as in a short traversal video.
To address these challenges, we propose \projname, a novel zero-shot, open-vocabulary semantic navigation framework that operates using only a monocular camera. Leveraging powerful 3D foundation models, \projname\ eliminates the dependency on depth and pose while exhibiting strong VICL capability. 
By simply observing a short video of the target environment, the system can also significantly improve task efficiency without requiring architectural modifications or task-specific retraining.
The framework integrates several key components: keyframe-based 3D reconstruction, semantic point cloud generation, vision-language model (VLM)-driven exploration value estimation, high-level adaptive waypoint selection, and low-level action execution. Experiments on the HM3D benchmark and real-world environments demonstrate that \projname\ achieves competitive performance in terms of navigation success rate and exploration efficiency, while showing superior VICL adaptability, with no previous 3D mapping of the environment required.
\end{abstract}

\section{INTRODUCTION}

\begin{figure}[!t]
\centering
\includegraphics[width=0.85\linewidth]{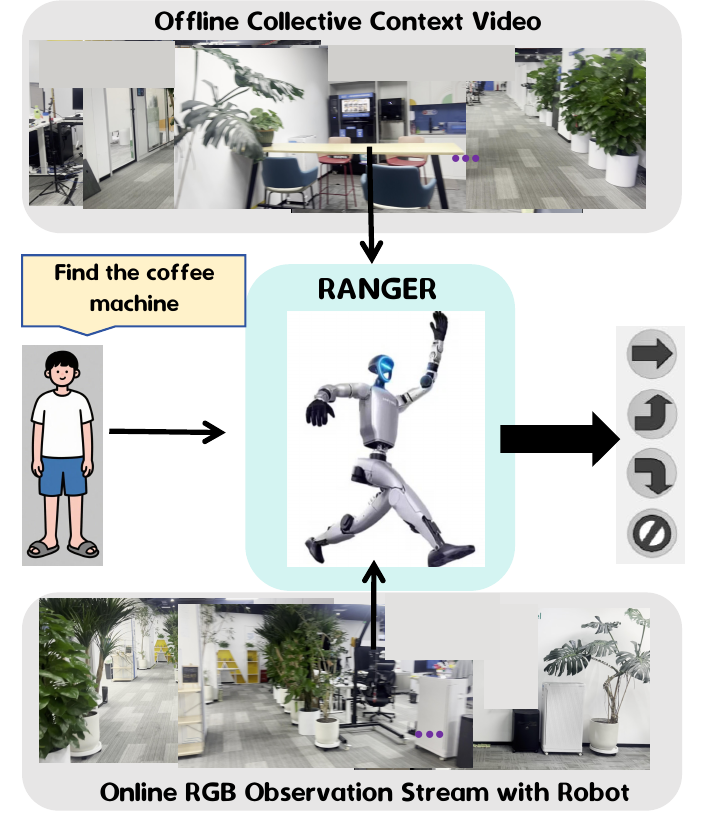}
\caption{\projname\ Workflow: Given an offline video of a new environment captured with a camera (optionally)  and an online RGB observation stream from the robot, \projname\ efficiently adapts to the new environment and completes navigation tasks based on human instructions.}
\label{fig:intro}
\vspace{-0.6cm}
\end{figure}

Efficient visual target localization and autonomous navigation in complex environments are core capabilities of embodied agents, with broad applications in service robotics, human–robot interaction, and autonomous monitoring systems. With the recent development of multimodal foundation models, including large language models (LLMs) (\eg \cite{achiam2023gpt, liu2023visual, bai2023qwen, grattafiori2024llama, team2023gemini,chen2025recoverable}) and open-vocabulary vision models (\eg \cite{liu2024grounding, kirillov2023segment, cheng2024yolo}), the performance of zero-shot object navigation (\eg \cite{kuang2024openfmnav, gadre2023cows, yu2023l3mvn, wu2024Voronav, yokoyama2024vlfm, yin2025unigoal}) has significantly improved. 
Such methods enable robots to be capable of navigating towards arbitrary target objects without task-specific fine-tuning.
Specifically, they leverage open-vocabulary perception to recognize arbitrary targets and semantic reasoning to guide navigation. 
This zero-shot paradigm enables immediate deployment in unseen environments without the need for environment-specific re-training.

However, existing approaches still face major limitations. First, most methods rely heavily on depth sensing and precise pose estimation, usually involving global mapping/reconstruction, which are typically feasible only in simulation or with specialized hardware. 
In real-world scenarios, sensor noise, calibration errors, data collection, and hardware costs substantially restrict their practicality, similarly to indoor localization scenarios~\cite{shu2019incrementally}. 
Second, current systems have yet to better leverage contextual knowledge available in everyday video streams, which can provide valuable priors for efficient exploration and rapid adaptation to novel environments.

This naturally raises a fundamental question: \textit{Can a navigation framework be built relying solely on RGB input?} Such a framework would be inherently low-cost, easy to deploy, and capable of directly utilizing videos captured from arbitrary devices for rapid adaptation to new environments. Notably, recent data-driven methods such as MASt3R~\cite{leroy2024grounding}, Dust3R~\cite{wang2024dust3r}, Fast3R~\cite{yang2025fast3r}, and VGGT~\cite{wang2025vggt} have demonstrated remarkable performance and generalization in 3D reconstruction tasks, opening new opportunities to achieve efficient navigation with RGB-only input.

In this paper, we introduce \projnamebold: a zero-shot, open-vocabulary semantic navigation framework based on monocular RGB input. Our approach requires neither depth sensors nor precise robot pose estimation, yet enables semantic-driven exploration and navigation in unseen environments.
Furthermore, \projname\ can adapt to a new environment using only a short offline video to extract rich geometric and semantic priors, significantly reducing exploration costs and enhancing navigation efficiency.
Importantly, our method does not require any task-specific fine-tuning during navigation (see Fig.~\ref{fig:intro}).
Specifically, \projname\ is built upon four key components: (1) online 3D reconstruction and localization from RGB-only input, (2) cross-frame semantic point cloud fusion, (3) high-level waypoint selection guided by vision–language models (VLMs), and (4) low-level action execution. At its core lies a \textit{keyframe-based memory bank}, which encodes geometric, semantic, and exploration-value information of the environment. This memory is continuously curated and optimized in the backend to reduce errors and improve system robustness.

We conduct systematic evaluations on the large-scale HM3D dataset, demonstrating that \projname\ achieves competitive performance without access to depth sensing,  precise poses, or global maps. In addition, we validate the practicality and robustness of our system in real-world environments such as meeting rooms and offices with a humanoid robot deployment. 

Our contributions are as follows: (1) A zero-shot semantic navigation framework (\projname) that operates solely on monocular RGB input, enabling rapid adaptation via offline videos. (2) A novel system architecture integrating RGB-based 3D reconstruction with VLM-guided hierarchical planning through a dynamic keyframe-based memory bank. (3) Extensive evaluations on HM3D and real-world humanoid robots, showing competitive performance and superior VICL adaptability.
\section{Related Work}

\textbf{Object Goal Navigation.} 
Current object-goal navigation methods include: (1) Learning-based approaches that train navigation policies over visual features extracted by pre-trained encoders via extensive imitation learning (\eg~\cite{ramrakhya2023pirlnav,zhang2024navid,batra2020objectnav,ehsani2024spoc,chen2023object,huang2023embodied,yu2025c}).
However, these methods often overfit to simulations and are restricted to predefined object categories, limiting open-world generalization.
(2) Approaches that focus on Zero-Shot Object Navigation (\eg~\cite{kuang2024openfmnav, gadre2023cows, yokoyama2024vlfm, wu2024Voronav,yin2025unigoal, yuan2025being, yu2024vln}). These methods explicitly construct environmental maps and utilize VLMs~\cite{ achiam2023gpt, bai2023qwen} or LLMs~\cite{grattafiori2024llama, team2023gemini} to select the most valuable frontier or waypoint for exploration. For instance, in Cow~\cite{gadre2023cows}, the robot explores the nearest frontier point until the target is detected using CLIP features~\cite{radford2021learning} and open-vocabulary object detectors~\cite{caron2021emerging}. ESC~\cite{zhou2023esc}, L3MVN~\cite{yu2023l3mvn}, and Voronav leverage LLMs for reasoning and decision-making to enhance performance. VLFM~\cite{yokoyama2024vlfm} employs a VLM to build semantic values on the map based on first-person observations and textual prompts, selecting the highest-scoring frontier. 
Their reliance on depth information, precise pose estimation, and task-specific fine-tuning often limits their performance in real-world scenarios, where depth data may be inaccurate or unavailable. These methods also hinder the ability to perform contextual learning, making it challenging for robots to leverage freely available monocular videos to improve exploration efficiency and adapt to new environments.

\textbf{Monocular 3D Reconstruction.}
Monocular 3D reconstruction methods have made notable progress recently in enabling the building of 3D maps from single camera inputs. Techniques such as MASt3R~\cite{leroy2024grounding}, Dust3R~\cite{wang2024dust3r}, Fast3R~\cite{yang2025fast3r}, VGGT~\cite{wang2025vggt}, MASt3R-SLAM~\cite{murai2025mast3r-slam}, and VGGT-SLAM~\cite{vggt-slam} have contributed significantly to this field. However, these methods often rely on passive reconstruction, where the trajectory is pre-collected and does not involve active perception or exploration. This restricts the system’s ability to adapt to dynamic environments and limits its exploration capabilities. In contrast, our approach tightly integrates incremental monocular 3D reconstruction with active semantic exploration to enable autonomous navigation.

\textbf{In-Context Learning.}
In-context learning (ICL)~\cite{brown2020language} allows models to perform tasks by leveraging a few examples provided in the input context, without requiring task-specific retraining. 
Recent work has extended this idea to robotics, where robots are trained to perform navigation tasks based on contextual information, such as video or textual prompts. For example, several works have used context from videos or previous maps to train navigation models \cite{zhou2025nolo}. However, these methods often require the target object to appear in the context, which severely limits their practical application, especially when the object is not initially visible.
In contrast, our approach does not require the target object to appear, even in the collected offline video. Instead, 
our approach enables robots to adaptively decide whether to continue exploring or exploit the knowledge contained in the offline video, remaining fully functional even without any prior visual context. This adaptive approach makes \projname\ more versatile, allowing it to handle a broader range of navigation scenarios.

\begin{figure*}[htbp]
    \centering
    \includegraphics[width=0.95\linewidth]{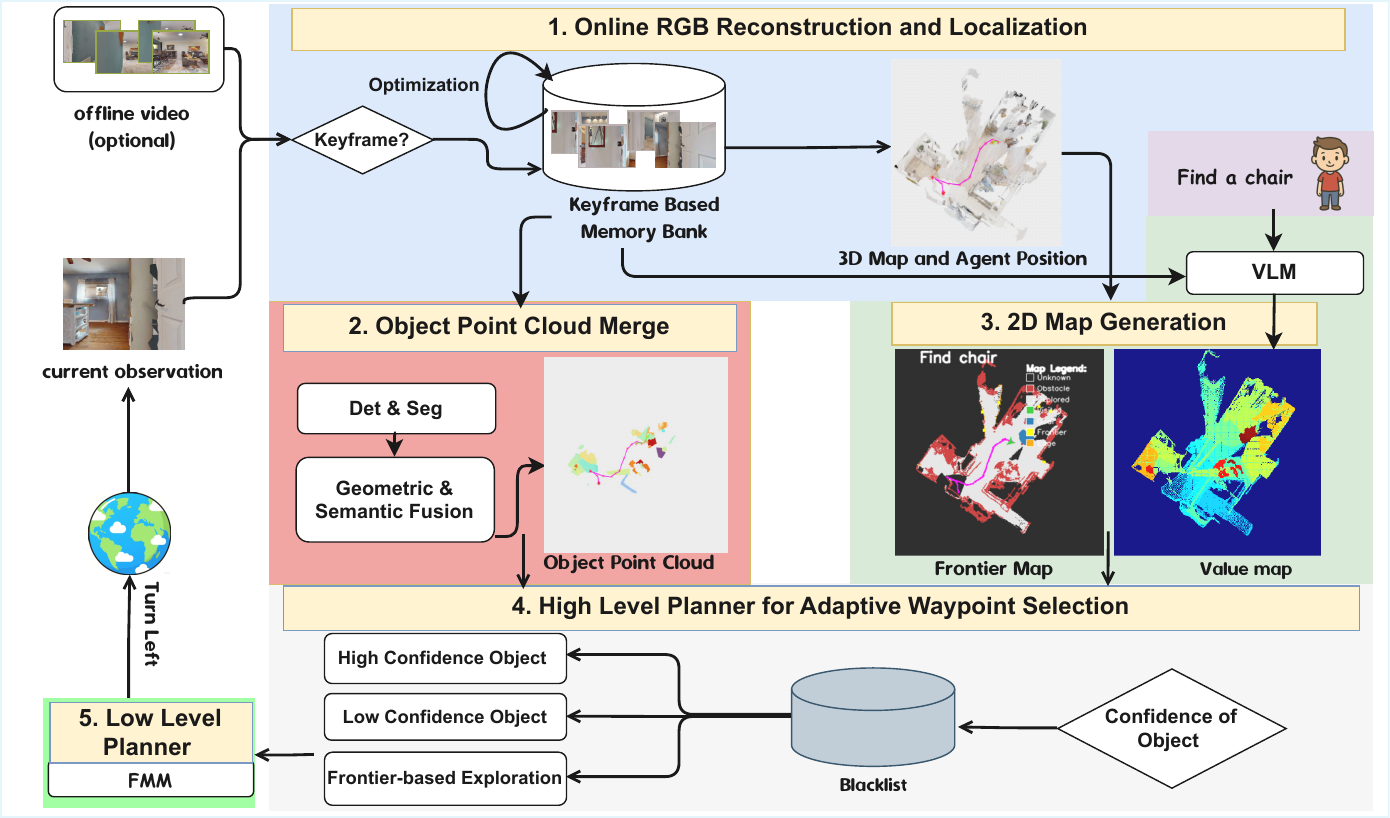}
    \caption{Overview of \projname: a unified architecture for zero-shot RGB-only navigation that builds geometric and semantic memory via a keyframe-based memory bank, which subsequently guides the agent’s exploration.}
\label{fig:model_diagram}
\vspace{-0.55cm}
\end{figure*}

\section{Problem Formulation}

We consider two semantic navigation tasks under a \textbf{zero-shot} setting, where the agent must locate a target object in a novel environment without task-specific training. These tasks differ in the availability of contextual information: the first relies solely on online RGB observations, while the second leverages an offline traversal video as prior context.

\textbf{RGB Navigation.}
In the standard \textit{ObjectNav} task with RGB-only inputs, an agent must navigate an unseen environment to approach a target object specified by a human-provided instruction. 
We denote the set of object categories as $\mathcal{C} = \{c_1, \dots, c_M\}$ and the set of scenes as $\mathcal{S} = \{s_1, \dots, s_N\}$. 
\footnote{Each scene $s_i \in \mathcal{S}$ represents a unique physical environment (e.g., a specific house layout) used to evaluate the agent's navigation performance.}
Each episode is defined by a tuple $\mathcal{T}_i = \langle s_i, c_i, p_i \rangle$, where the agent starts from an initial pose $p_i$ in scene $s_i$ with target category $c_i \in \mathcal{C}$. 
At each time step $t$, the agent receives an RGB observation $o_t \in \mathbb{R}^{H \times W \times 3}$ and the target category $c_i$, without access to depth or pose information. The action space $\mathcal{A}$ consists of six discrete actions: \textit{move\_forward}, \textit{turn\_left}, \textit{turn\_right}, \textit{look\_up}, \textit{look\_down}, and \textit{stop}. 
An episode is considered successful if the agent executes a \textit{stop} action when the geodesic distance to the target is below a predefined threshold, with a trajectory length within the step limit.

\textbf{Contextual RGB Navigation.}
In the contextual variant, the agent is additionally provided with an offline traversal video $V_{s_i}$ of a human exploring the target environment $s_i$. The agent must then navigate the same environment $s_i$ to approach the target object by combining prior visual context from $V_{s_i}$ with real-time RGB observations. Compared to the RGB-only setting, this formulation enables the agent to reduce redundant exploration and achieve more efficient semantic navigation in unseen environments.

\section{Methodology}

\vspace{-0.07cm}
\subsection{Overview of the Process}

As illustrated in Fig.~\ref{fig:model_diagram}, \projname\ employs a unified architecture that seamlessly handles both \textit{zero-shot RGB-only navigation} and \textit{video-context-augmented navigation} without requiring any architectural modifications or retraining. The system maintains a dynamic keyframe bank, which serves as a joint geometric and semantic memory of the environment. Upon receiving each new RGB observation, the backend refines the agent's pose in real-time and performs loop closure by optimizing relative poses against the stored keyframes. This process incrementally constructs a dense 3D  map, enabling metric self-localization. 
Meanwhile, a semantic module aggregates object features via CLIP~\cite{radford2021learning} and GroundingDINO~\cite{liu2024grounding} into a semantic point cloud, where each point is annotated with category probabilities and spatial context.
To facilitate efficient navigation, we project the 3D reconstruction into a 2D occupancy map. Based on this representation, a hierarchical planner generates high-level waypoints by jointly evaluating the target's semantic relevance to the textual goal, along with its spatial and perceptual confidence.
This perception-planning-action loop executes continuously until the agent reaches the target.
\subsection{Online RGB Reconstruction and Localization}

To enable metric-aware navigation from monocular RGB inputs, we integrate \textbf{MASt3R-SLAM}~\cite{murai2025mast3r-slam}, a state-of-the-art real-time dense SLAM system based on the MASt3R two-view 3D reconstruction prior. Its open-source implementation is directly incorporated into \projname\ as a plug-and-play geometric backbone.

\textbf{Keyframe-Based Reconstruction.} To build the dense scene representation, the pipeline dynamically registers keyframes when the number of valid matches falls below a specific threshold  $\omega_k$ . For each frame, the MASt3R network directly predicts dense pointmaps. The system continuously fuses them into a canonical keyframe pointmap using a running weighted average filter. 
This sequential filtering seamlessly merges multi-view information to maintain a coherent 3D structure.

\textbf{Localization.}
To continuously estimate the camera pose, the system first establishes fast pixel correspondences between the incoming frame and the active keyframe using an efficient iterative projective matching scheme. Based on these matches, the relative pose is then solved by minimizing the directional ray error. 
By formulating the problem purely around ray geometry, this approach inherently bypasses the need for specific camera models. Consequently, it remains highly robust to structural inconsistencies and seamlessly handles uncalibrated video sequences, making it ideal for real-world deployment.

\textbf{Global Consistency via Backend Optimization.} To mitigate drift, MASt3R-SLAM constructs a pose graph connected by sequential tracking and loop-closure edges. Loop candidates are efficiently retrieved via the ASMK framework~\cite{tolias2013aggregate} on encoded visual features and validated through geometric matching. The system then enforces long-term consistency across all poses and dense geometry by jointly minimizing the directional ray errors across the entire graph.

\textbf{Real-Scale Reconstruction.} 
Since monocular reconstruction lacks absolute scale, we utilize the robot's camera height $h_{\text{real}}$ as a geometric prior. We select the lowest 20\% of points along the vertical axis as candidate ground points and use RANSAC to fit a ground plane $\mathbf{n}^\top \mathbf{x} + d = 0$, where $\mathbf{n} = [a, b, c]^\top$ is the plane normal. The estimated camera height in the reconstructed space is computed as $h_{\text{est}} = \frac{|d|}{\|\mathbf{n}\|}$. 
Finally, we derive the global scale factor $s = h_{\text{real}} / h_{\text{est}}$ to align the 3D map with the physical environment.

\subsection{Object Point Cloud Merge}
We first employ Grounding DINO as an open-vocabulary detector and combine it with Mobile SAM~\cite{kirillov2023segment} to extract object masks within the detection boxes of each keyframe stored in the memory bank. Then, using the keyframe confidence masks provided by MASt3R-SLAM, we filter these object regions to obtain high-confidence object point cloud representations in 3D space. For each semantic mask, we further crop the corresponding image region and extract visual features with CLIP. This process ultimately generates a semantic point cloud along with class confidence scores and target category information.

\textbf{Object Association.}
For each newly detected object  $j$, we compute both geometric and semantic similarity to existing objects. Geometric similarity $S_{\text{geo}}$ is based on the Intersection over Union (IoU) of the point clouds, while visual similarity $S_{\text{vis}}$ is determined by comparing the CLIP semantic features of the newly detected object with those of existing objects. The overall similarity score is a weighted sum:
\begin{equation}
    s(i, j) = w_1 S_{\text{vis}}(i, j) + w_2 S_{\text{geo}}(i, j),
\end{equation}
where $w_1$ and $w_2$ are weights. If the similarity score falls below a predefined threshold $\tau$, a new object is initialized. Otherwise, the new detection is associated with the existing object exhibiting the highest similarity.

\textbf{Object Fusion.} For associated objects, their point clouds are merged, and their features are updated via a weighted average based on detection frequency. Additionally, the final confidence of the fused object is determined by the maximum confidence score across all its observations.

\subsection{2D Map Generation}
\textbf{Frontier Map.} After constructing a unified 3D point cloud, we project the points onto a 2D plane to create obstacle and exploration maps. Points above the floor are projected onto the obstacle map, while all points contribute to the exploration map. Frontiers are generated by detecting the boundaries between explored and unexplored regions. As the robot explores, the locations of frontiers dynamically change until the environment is fully explored.

\textbf{Value Map.} To guide the agent's exploration, we construct a 2D value map by integrating both object-level and image-level semantic relevance. For the fused objects and each keyframe in the memory bank,  
we compute the cosine similarity between their CLIP embeddings $\mathbf{f}$ and the target instruction embedding $\mathbf{g}$, i.e., $\frac{\mathbf{f} \cdot \mathbf{g}}{\|\mathbf{f}\| \|\mathbf{g}\|}$, denoted as matching scores $s_{\text{obj}}$ and $s_{\text{img}}$ respectively.
The associated high-confidence 3D point clouds are then projected onto a 2D grid. The final value at each grid cell $(i, j)$ is determined by the maximum relevance score from both sources, denoted as $V(i, j) = \max(s_{\text{obj}}, s_{\text{img}})$.

\subsection{High-Level Planner for Adaptive Waypoint Selection}
To ensure robust navigation under perceptual uncertainty, we design a high-level planner that dynamically selects waypoints by integrating \textit{semantic relevance}, \textit{traversability}, and \textit{candidate confidence}.
Operating on dense 2D maps and the current pose $T_t$ provided by MASt3R-SLAM, the planner evaluates a candidate set $\mathcal{W}_t = \mathcal{O}_t \cup \mathcal{F}_t$, where $\mathcal{O}_t$ represents object proposals and $\mathcal{F}_t$ denotes frontier points. The selection follows a \textit{hierarchical priority}: (1) objects exceeding $\sigma_{obj}$ are final targets; (2) low-confidence objects serve as temporary waypoints for guided exploration; (3) otherwise, the frontier in $\mathcal{F}_t$ with the highest 2D value map response is selected. To ensure \textit{geometric feasibility}, all candidates must be navigable via the obstacle map. 
Crucially, a dynamic blacklist $\mathcal{B}_t$ records two failure modes to prevent repetitive errors: (i) \textit{unreachable locations} that remain inaccessible beyond the maximum step limit, and (ii) \textit{perceptual false positives} where low-confidence candidates fail to meet the verification criteria despite exceeding $\tau_{\text{obs}}$ cumulative observations.

\subsection{Low-Level Planner for Path Planning}
We use the Fast Marching Method (FMM)~\cite{sethian1996fast} as the local planner to generate a path toward the goal. At each step, the robot executes motion commands toward a local subgoal, while the map and target are updated in real time, enabling adaptive, closed-loop navigation in partially observed environments.

\begin{figure*}[htbp]
    \centering
    \includegraphics[width=1.0\linewidth]{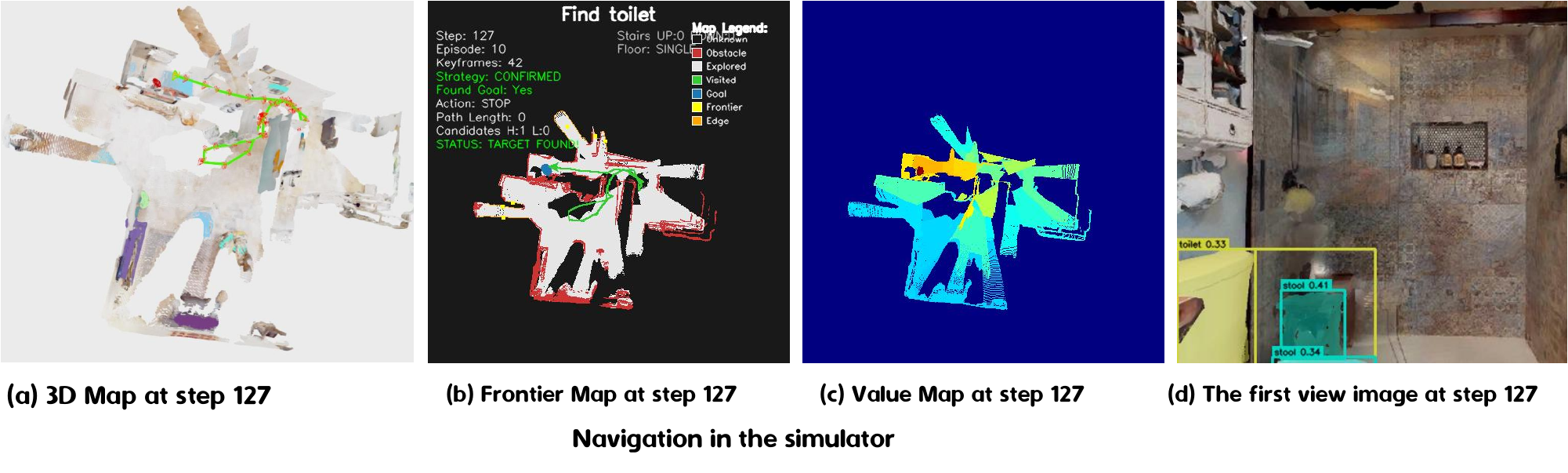}
    \caption{Visualization of the navigation process in a simulated environment with task instruction: "Find the toilet". 
    (a) Reconstructed 3D map and the agent's trajectory at step 127. The red cones represent keyframe locations, while the green cone indicates the agent's current position. The green line traces the agent's movement, and the red point cloud represents the toilet object. 
    (b) The Frontier Map showing the robot's exploration trajectory. The blue area highlights the target object region, red areas indicate obstacles, and yellow dots represent frontier points.
    (c) The Value Map, where red regions correspond to high-value areas.
    (d) First-person observation.}
\label{fig:rgbnav}
\vspace{-0.33cm}
\end{figure*}

\section{Experiments}
We evaluate \projname\ on object goal navigation tasks across both simulated and real-world environments, testing its navigation performance in completely unseen settings, as well as when provided with a contextual video.

\subsection{Simulation-based Experiment Setup}

\textbf{Navigation in unseen environments.}  
We evaluate our approach on the Habitat HM3Dv2 dataset (HM3D-Semantics-v0.2~\cite{yadav2023habitat}, released in the 2023 Habitat Challenge), which includes six goal categories. 
For our experiments, we select 10 single-floor scenes comprising 279 episodes in total. 
In each episode, the agent’s starting position is randomly sampled to assess navigation performance in unseen environments.
The \textit{move forward} action moves the agent by 25 cm, while \textit{turn} actions adjust orientation by 30°.
Considering the trade-off between system storage and performance, we set the maximum episode length to 300 steps.

\textbf{Navigation with contextual video.}  
We use the same 10 scenes and manually collect one video trajectory per scene by teleoperating the robot in the simulator. For evaluation, the agent is initialized near the endpoint of the collected video to ensure partial overlap between the agent’s initial observation and the video frames, which facilitates re-localization. For each scene, we randomly sample 2 or 3 starting positions near the video endpoint, resulting in 100 episodes in total across 6 object categories.

\subsection{Metrics}

We use navigation success rate (SR) and success rate weighted by navigation path length (SPL) as evaluation metrics. SR represents the percentage of successful episodes out of the total episodes. SPL measures the efficiency of reaching the goal in addition to the success rate.

\subsection{Implementation Details}

In Habitat, the agent uses a 640$\times$480 RGB camera mounted at 0.88m height.
For semantic segmentation, we leverage Grounding DINO~\cite{liu2024grounding}  to generate class-agnostic object proposals and MobileSAM~\cite{kirillov2023segment} to produce pixel-accurate masks. Semantic confidence scores for detected objects are computed using CLIP~\cite{radford2021learning}, enabling zero-shot semantic grounding.
Our system employs consistent hyper-parameters: keyframe selection threshold $\omega_k = 0.333$, loop-closure detection threshold $\omega_l = 0.1$, and robustness regularization weight $\omega_r = 0.005$. To ensure reliable relocalization, we enforce a stricter geometric consistency check when attaching the current frame to the pose graph: the ratio of inlier feature matches must exceed $0.3$. In addition, the confidence threshold for the point cloud is set to $1.9$.

\section{Results}

\subsection{Navigation in Unseen Environments}
Previously, zero-shot object navigation heavily relied on accurate pose and depth information provided by simulators or auxiliary hardware sensors. To evaluate our approach in more realistic settings, we compare with L3MVN~\cite{yu2023l3mvn}, a representative work that utilizes LLMs for efficient frontier selection and requires privileged inputs (depth and pose). As detailed in Table~\ref{tab:rgbnav_metrics}, L3MVN (300) and L3MVN (500) represent performance under step limits of 300 and 500, respectively, both leveraging RGB, depth, and pose data. In contrast, \projname\ operates solely on RGB input, without any privileged sensing.

\begin{table}[htbp]
  \centering
  \caption{Navigation performance in unseen environments.}
  \label{tab:rgbnav_metrics}
  \begin{tabular}{l c c c c}  
    \toprule
    Methods & Sensors & SR & SPL & Avg Steps \\
    \midrule
    L3MVN (300)     & RGB, Depth, Pose & 39.4 & 15.7 & 147.62 \\
    L3MVN (500)     & RGB, Depth, Pose & 42.3 & 16.3 & 171.06 \\
    \midrule
    \projname\ (300) & RGB & 42.7 & 17.8 & 172.80 \\
    \bottomrule
  \end{tabular}
\end{table}

As presented in Table~\ref{tab:rgbnav_metrics}, \projname\ achieves an SR of 42.7\% and an SPL of 17.8\% under the 300-step limit. This performance outperforms L3MVN (300) by 3.3 percentage points in SR. Notably, \projname\ even surpasses L3MVN (500), which benefits from both an extended step limit and privileged sensor inputs. These compelling results underscore that integrating contextual video priors enables competitive navigation performance in previously unseen environments using only standard RGB input. Figure~\ref{fig:rgbnav} visually depicts a navigation instance: the agent successfully reaches the \textit{toilet} at step 127, with 42 frames stored in its keyframe-based memory bank. The highlighted region in Figure~\ref{fig:rgbnav} (c) further confirms the strong spatial consistency between the agent's memory and the observed environment, illustrating effective visual grounding.

\subsection{Navigation with a Contextual Video}
\vspace{-0.2cm}
\begin{figure*}[htbp]
    \centering
    \includegraphics[width=1.0\linewidth]{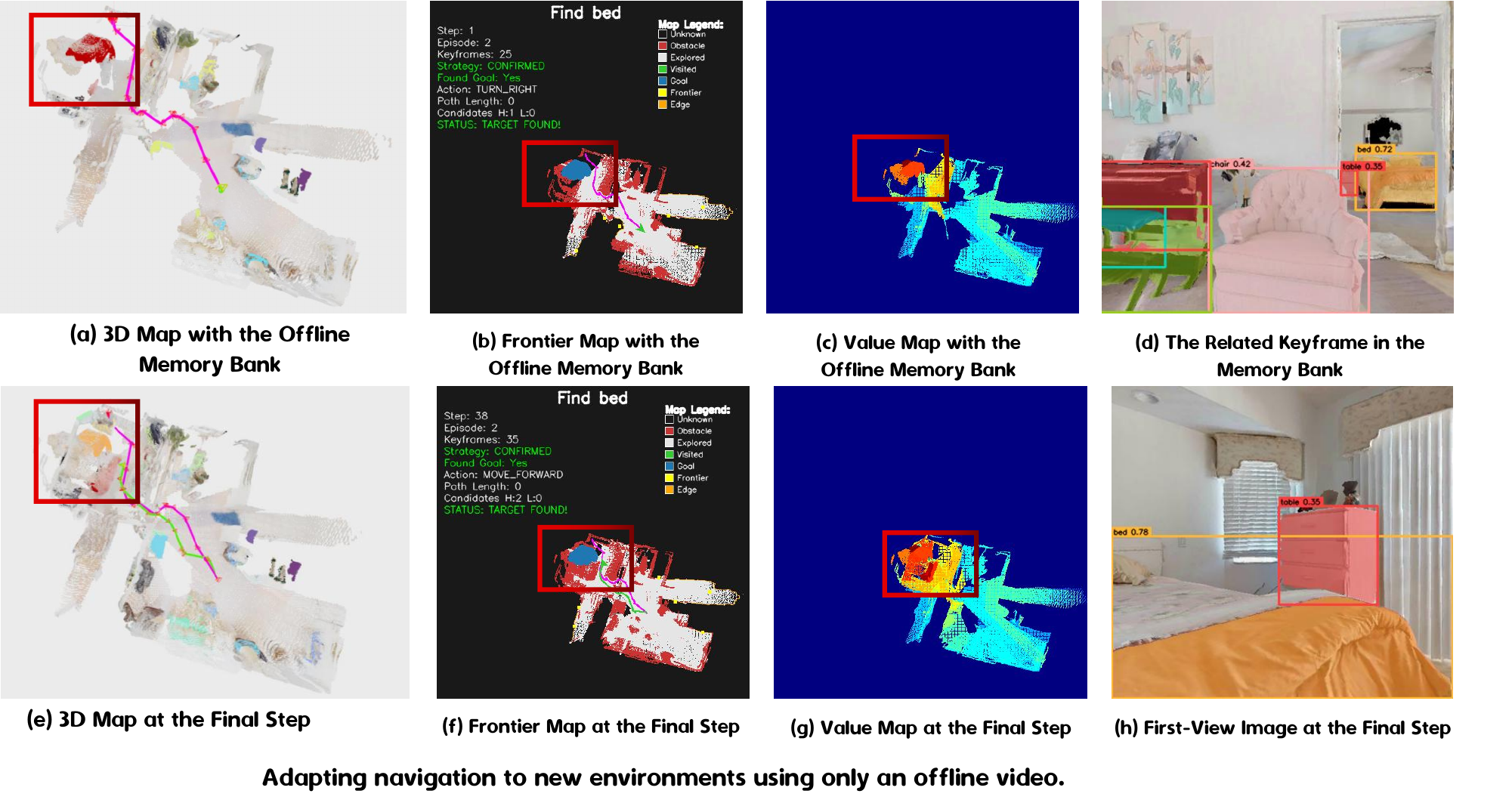}
    \caption{
Contextual adaptation for navigation in a new environment. \projname\ can leverage an offline video to adapt and locate the target object "bed". (a-d) Initial state with offline memory; (e-h) Final navigation state. The system successfully found the target in 38 steps.}
\label{fig:context_rgbnav}
\vspace{-0.3cm}
\end{figure*}

\begin{table}[htbp]
  \centering
  \caption{Performance w/ \& w/o contextual video.}
  \begin{tabular}{l c c c c}
    \toprule
    Methods & SR & SPL & DTG & Avg Steps \\
    \midrule
    w/o video & 44.0 & 16.4 & 2.933  & 171.36   \\
    w/  video & 58.0 & 30.2 & 2.381 & 110.22 \\
    \bottomrule
  \end{tabular}
  \vspace{-0.30cm}
  \label{tab:contextual_video}
\end{table}

Table~\ref{tab:contextual_video} presents the navigation performance with and without contextual video. 
As shown in Table II, incorporating an offline traversal video significantly improves performance, boosting SR ($44.0\% \rightarrow 58.0\%$) and SPL ($16.4 \rightarrow 30.2$), while reducing average steps ($171.4 \rightarrow 110.2$).
These results confirm that contextual videos effectively reduce redundant exploration and guide the agent toward more efficient navigation.  
In our framework, offline videos serve as an \textit{environmental preview}. 
By constructing a keyframe-based memory bank, the system acquires prior knowledge of the scene before navigation, thereby reducing redundant exploration. 
This mechanism mimics the way humans navigate more efficiently in a new environment after watching a walkthrough video. 
In practice, we observe that providing such contextual information enables the agent to significantly reduce unnecessary exploration and complete target-finding tasks more efficiently. 
As illustrated in Fig.~\ref{fig:context_rgbnav}, when searching for a \textit{bed}, \projname\ first builds a keyframe memory bank from the offline video and applies semantic retrieval to identify the target object within the video. 
The object is then transformed into a point cloud representation, which is integrated with both the offline video and real-time observations to construct an environment map. 
Finally, the system plans a collision-free path that directly leads the agent to the target location.

\subsection{Ablation Studies}
\begin{figure}[htbp]
    \centering
    \includegraphics[width=1.03\linewidth]{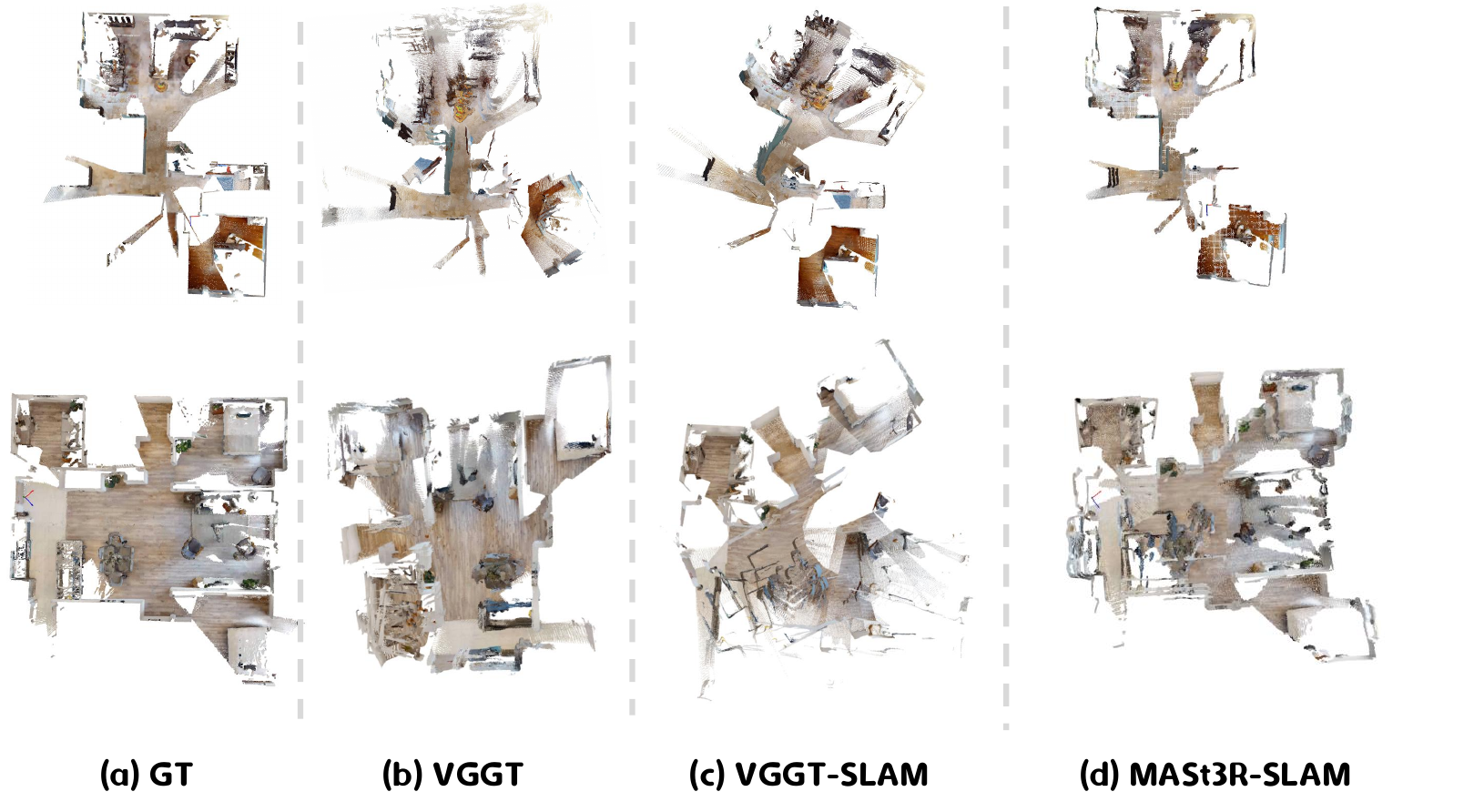}
    \caption{
    Comparison of 3D maps reconstructed using different methods. Figure (a) shows the result using real depth and pose data. Figures (b), (c), and (d) present the reconstruction results of VGGT, VGGT-SLAM, and MASt3R-SLAM, respectively, all relying solely on RGB images.
    }
\label{fig:recons}
\vspace{-0.4cm}
\end{figure}

\textbf{Reconstruction and Localization Accuracy}  
As shown in Fig.~\ref{fig:recons}, we compare the latest 3D reconstruction methods (e.g., VGGT, VGGT-SLAM, and MASt3R-SLAM) with the ground truth.
The video trajectories are collected by a manually teleoperated agent navigating within the environment.  
The results demonstrate that VGGT suffers from insufficient detail modeling and, due to its requirement to process the entire video sequence at once, fails to meet the real-time demands of navigation tasks. In contrast, VGGT-SLAM constructs dense 3D maps and performs localization by incrementally aligning submaps generated by VGGT. However, in long-sequence scenarios, the submap merging process often introduces significant alignment errors, limiting its effectiveness.  
MASt3R-SLAM exhibits stronger robustness under the same conditions. It consistently achieves higher reconstruction quality and global consistency across trajectories of varying lengths, effectively mitigating drift and distortion issues commonly observed in long-sequence stitching. Moreover, it supports real-time operation, making it suitable for navigation applications.

\begin{figure*}[htbp]
    \centering
    \includegraphics[width=1.0\linewidth]{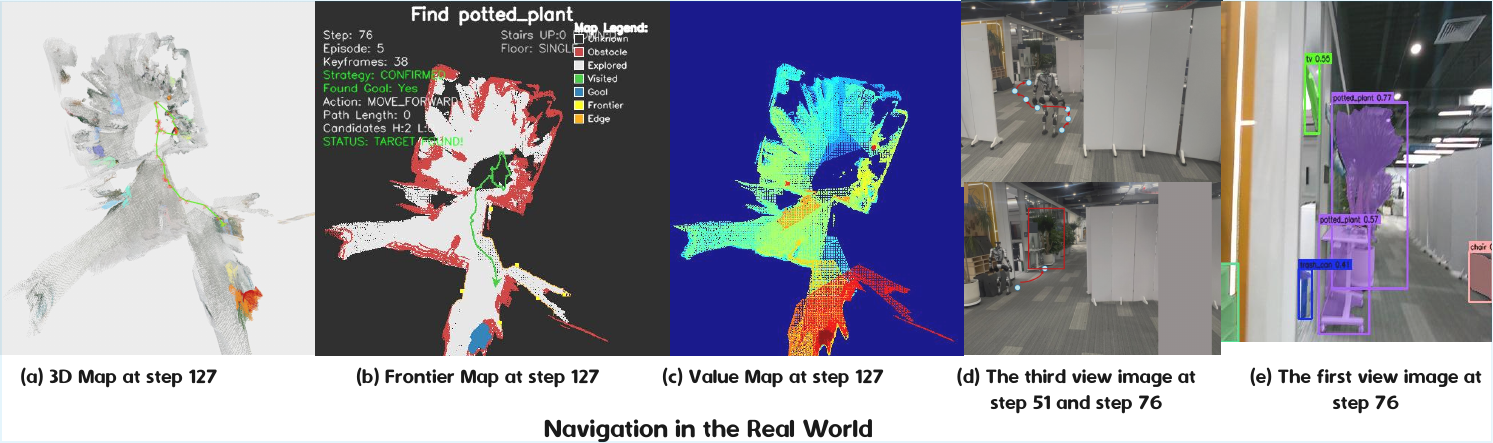}
    \caption{
       Real-world navigation with \projname\ on a Unitree G1 humanoid robot. Task: "Find potted plant.".}
\label{fig:real_navigation}
\vspace{-0.4cm}
\end{figure*}

\textbf{Component Analysis.} 
Table~\ref{tab:component_analysis} summarizes the ablation results for two key components: the high-level planner and the object detector. 
When the high-level planner is removed, the SR decreases from 42.7\% to 39.8\%, and the average number of steps increases from 172.80 to 184.43.
This confirms the critical role of the planner in efficiently guiding the agent toward the target.
For the detector comparison, YOLO-World~\cite{cheng2024yolo} and DINO yield relatively close results, with DINO showing a slight advantage (SR = 42.7\%, SPL = 17.8\% vs. 42.3\%, 17.1\%). 
This suggests that while replacing the detector has only a modest impact on overall accuracy, stronger visual grounding from DINO provides more consistent navigation performance.
\begin{table}[htbp]
  \centering
  \caption{Impact of model components.}
  \begin{tabular}{l c c c c c}
    \toprule
    component &  SR & SPL & DTG & Avg Steps \\
    \midrule
      w/o High Level Planner  &   39.8 & 17.6 & 3.552 & 184.43 \\
   \midrule
    YOLO-World &  42.3 & 17.1 & 3.573 & 178.06 \\
    DINO & 42.7 & 17.8 & 3.366& 172.80 \\
    \bottomrule
  \end{tabular}
  \label{tab:component_analysis}
\end{table}
\vspace{-0.15cm}

\textbf{Effect of Step Limits.}
As the step limit increases, the memory bank must maintain a growing number of keyframe point clouds, poses, semantic segmentation, and their associated values, potentially leading to substantial storage and computation overhead. 
Based on this consideration, we evaluate \projname\ under different maximum step limits. 
As shown in Table~\ref{tab:step_limits}, the success rate (SR) improves steadily with larger step budgets, rising from 6.1\% at 50 steps to 42.7\% at 300 steps. 
This indicates that a longer execution horizon provides the agent with more opportunities for exploration. 
However, the improvement becomes marginal beyond 250 steps; therefore, we set the maximum step limit to 300 in our experiments.
\begin{table}[htbp]
  \centering
  \caption{SR under different maximum step limits.}
  \begin{tabular}{l c c c c c}
    \toprule
    Max Steps & 50 & 100 & 200 & 250 & 300 \\
    \midrule
    SR        & 6.1 & 22.2 & 36.6 & 40.1 & 42.7 \\
    \bottomrule
  \end{tabular}
  \label{tab:step_limits}
   \vspace{-0.35cm}
\end{table}

\subsection{Navigation in the Real World}

\textbf{Real-World Experiment Setup.} 
We validate the practicality of \projname\ through real-world experiments on a Unitree G1 humanoid robot, equipped with an Intel RealSense D455 RGB-D camera mounted at a height of 1.29\,m on a custom 2 DoF head. 
Importantly, \textit{only the RGB stream is used} for sensing and navigation, consistent with our depth-free assumption. 
All system components—including MASt3R-SLAM for online 3D reconstruction, CLIP, and Grounding DINO for semantic grounding, and MobileSAM for instance-aware segmentation—are deployed on a mobile workstation with an NVIDIA RTX 4090 Laptop GPU and an Intel Core i9-13900HX CPU, achieving real-time inference at approximately 1 Hz.
The experiments were carried out in diverse indoor environments, including laboratories and meeting rooms, where the robot was instructed to navigate toward semantically defined targets. 

\textbf{Visualization.} 
Representative qualitative results are shown in Fig.~\ref{fig:real_navigation}. 
In the example task, the user instructs the robot to locate a \textit{plant}. 
After completing exploration of the first room (step 51), the robot autonomously selects a high-value frontier region and transitions to an adjacent room, successfully detecting the target at step 76. 
Notably, the high-value region highlighted in Fig.~\ref{fig:real_navigation} (c) aligns closely with the actual target location shown in Fig.~\ref{fig:real_navigation} (b), demonstrating the effectiveness of semantic information in guiding exploration.

\section{Conclusions and Limitations}

In this work, we introduce \projname, a novel zero-shot visual target navigation framework relying solely on RGB input. \projname\ significantly enhances navigation efficiency without task-specific fine-tuning through its keyframe-based memory bank, back-end optimization, keyframe-driven reconstruction, and hierarchical planning strategy.
The system also demonstrates rapid adaptation and substantial performance gains with only a short video sequence of a new environment, offering a new perspective on zero-shot navigation.
Ultimately, this work presents a practical pathway toward real-world autonomous navigation and opens new directions for research focused purely on visual input.

\textbf{Limitations.} 
Despite the strong performance and contextual learning capability of \projname, several limitations remain. First, the continuous accumulation of keyframes increases storage overhead and may degrade system efficiency over time, highlighting the need for more effective memory management strategies. Second, reconstruction and localization remain possible bottlenecks for purely RGB-based navigation, as their robustness is still insufficient to handle overly complex or dynamic environments. Enhancing the reliability of these modules will be critical for advancing the practical deployment of such systems.

\section{ACKNOWLEDGEMENTS}
\small{
This work is supported by the National Natural Science Foundation of China (Grant No. 62441617).
}


\bibliographystyle{IEEEtran}
\bibliography{IEEEabrv,References}

@article{batra2020objectnav,
  title={Objectnav revisited: On evaluation of embodied agents navigating to objects},
  author={Batra, Dhruv and Gokaslan, Aaron and Kembhavi, Aniruddha and Maksymets, Oleksandr and Mottaghi, Roozbeh and Savva, Manolis and Toshev, Alexander and Wijmans, Erik},
  journal={arXiv preprint arXiv:2006.13171},
  year={2020}
}

@inproceedings{ramrakhya2023pirlnav,
  title={Pirlnav: Pretraining with imitation and rl finetuning for objectnav},
  author={Ramrakhya, Ram and Batra, Dhruv and Wijmans, Erik and Das, Abhishek},
  booktitle={CVPR},
  pages={17896--17906},
  year={2023}
}

@inproceedings{chen2023object,
  title={Object goal navigation with recursive implicit maps},
  author={Chen, Shizhe and Chabal, Thomas and Laptev, Ivan and Schmid, Cordelia},
  booktitle={IROS},
  pages={7089--7096},
  year={2023}
}

@article{zhang2024navid,
  title={Navid: Video-based vlm plans the next step for vision-and-language navigation},
  author={Zhang, Jiazhao and Wang, Kunyu and Xu, Rongtao and Zhou, Gengze and Hong, Yicong and Fang, Xiaomeng and Wu, Qi and Zhang, Zhizheng and Wang, He},
  journal={RSS},
  year={2024}
}

@inproceedings{ehsani2024spoc,
  title={Spoc: Imitating shortest paths in simulation enables effective navigation and manipulation in the real world},
  author={Ehsani, Kiana and Gupta, Tanmay and Hendrix, Rose and Salvador, Jordi and Weihs, Luca and others},
  booktitle={CVPR},
  pages={16238--16250},
  year={2024}
}

@inproceedings{kuang2024openfmnav,
  title={Openfmnav: Towards open-set zero-shot object navigation via vision-language foundation models},
  author={Kuang, Yuxuan and Lin, Hai and Jiang, Meng},
  booktitle={Findings of the Association for Computational Linguistics: NAACL 2024},
  pages={338--351},
  year={2024}
}

@inproceedings{gadre2023cows,
  title={Cows on pasture: Baselines and benchmarks for language-driven zero-shot object navigation},
  author={Gadre, Samir Yitzhak and Wortsman, Mitchell and Ilharco, Gabriel and Schmidt, Ludwig and Song, Shuran},
  booktitle={CVPR},
  pages={23171--23181},
  year={2023}
}

@inproceedings{yu2023l3mvn,
  title={L3mvn: Leveraging large language models for visual target navigation},
  author={Yu, Bangguo and Kasaei, Hamidreza and Cao, Ming},
  booktitle={IROS},
  pages={3554--3560},
  year={2023}
}

@article{wu2024voronav,
  title={Voronav: Voronoi-based zero-shot object navigation with large language model},
  author={Wu, Pengying and Mu, Yao and Wu, Bingxian and Hou, Yi and Ma, Ji and Zhang, Shanghang and Liu, Chang},
  journal={arXiv preprint arXiv:2401.02695},
  year={2024}
}

@inproceedings{yokoyama2024vlfm,
  title={Vlfm: Vision-language frontier maps for zero-shot semantic navigation},
  author={Yokoyama, Naoki and Ha, Sehoon and Batra, Dhruv and Wang, Jiuguang and Bucher, Bernadette},
  booktitle={ICRA},
  pages={42--48},
  year={2024}
}

@inproceedings{zhou2023esc,
  title={Esc: Exploration with soft commonsense constraints for zero-shot object navigation},
  author={Zhou, Kaiwen and Zheng, Kaizhi and Pryor, Connor and Shen, Yilin and Jin, Hongxia and Getoor, Lise and Wang, Xin Eric},
  booktitle={ICML},
  pages={42829--42842},
  year={2023}
}

@article{yin2025unigoal,
  title={UniGoal: Towards Universal Zero-shot Goal-oriented Navigation},
  author={Yin, Hang and Xu, Xiuwei and Zhao, Lingqing and Wang, Ziwei and Zhou, Jie and Lu, Jiwen},
  journal={CVPR},
  year={2025}
}

@article{huang2023embodied,
  title={An embodied generalist agent in 3d world},
  author={Huang, Jiangyong and Yong, Silong and Ma, Xiaojian and Linghu, Xiongkun and Li, Puhao and Wang, Yan and Li, Qing and Zhu, Song-Chun and Jia, Baoxiong and Huang, Siyuan},
  journal={arXiv preprint arXiv:2311.12871},
  year={2023}
}

@inproceedings{yadav2023habitat,
  title={Habitat-matterport 3d semantics dataset},
  author={Yadav, Karmesh and Ramrakhya, Ram and Ramakrishnan, Santhosh Kumar and Gervet, Theo and Turner, John and Gokaslan, Aaron and Maestre, Noah and others},
  booktitle={CVPR},
  pages={4927--4936},
  year={2023}
}

@inproceedings{liu2024grounding,
  title={Grounding dino: Marrying dino with grounded pre-training for open-set object detection},
  author={Liu, Shilong and Zeng, Zhaoyang and Ren, Tianhe and Li, Feng and Zhang, Hao and Yang, Jie and Jiang, Qing and Li, Chunyuan and Yang, Jianwei and Su, Hang and others},
  booktitle={ECCV},
  pages={38--55},
  year={2024}
}

@inproceedings{kirillov2023segment,
  title={Segment anything},
  author={Kirillov, Alexander and Mintun, Eric and Ravi, Nikhila and Mao, Hanzi and Rolland, Chloe and Gustafson, Laura and Xiao, Tete and Whitehead, Spencer and Berg, Alexander C and Lo, Wan-Yen and others},
  booktitle={ICCV},
  pages={4015--4026},
  year={2023}
}

@inproceedings{cheng2024yolo,
  title={Yolo-world: Real-time open-vocabulary object detection},
  author={Cheng, Tianheng and Song, Lin and Ge, Yixiao and Liu, Wenyu and Wang, Xinggang and Shan, Ying},
  booktitle={CVPR},
  pages={16901--16911},
  year={2024}
}

@article{yuan2025being,
  title={Being-0: A Humanoid Robotic Agent with Vision-Language Models and Modular Skills},
  author={Yuan, Haoqi and Bai, Yu and Fu, Yuhui and Zhou, Bohan and Feng, Yicheng and Xu, Xinrun and others},
  journal={arXiv preprint arXiv:2503.12533},
  year={2025}
}

@inproceedings{radford2021learning,
  title={Learning transferable visual models from natural language supervision},
  author={Radford, Alec and Kim, Jong Wook and Hallacy, Chris and Ramesh, Aditya and Goh, Gabriel and Agarwal, Sandhini and Sastry, Girish and Askell, Amanda and Mishkin, Pamela and Clark, Jack and others},
  booktitle={ICML},
  pages={8748--8763},
  year={2021}
}

@inproceedings{shu2019incrementally,
  title={Incrementally-deployable indoor navigation with automatic trace generation},
  author={Shu, Yuanchao and Li, Zhuqi and Karlsson, B{\"o}rje and Lin, Yiyong and Moscibroda, Thomas and Shin, Kang},
  booktitle={IEEE INFOCOM 2019-IEEE Conference on Computer Communications},
  pages={2395--2403},
  year={2019}
}

@article{achiam2023gpt,
  title={Gpt-4 technical report},
  author={Achiam, Josh and Adler, Steven and Agarwal, Sandhini and Ahmad, Lama and Akkaya, Ilge and Aleman, Florencia Leoni and Almeida, Diogo and Altenschmidt, Janko and Altman, Sam and Anadkat, Shyamal and others},
  journal={arXiv preprint arXiv:2303.08774},
  year={2023}
}

@article{liu2023visual,
  title={Visual instruction tuning},
  author={Liu, Haotian and Li, Chunyuan and Wu, Qingyang and Lee, Yong Jae},
  journal={NeurIPS},
  volume={36},
  pages={34892--34916},
  year={2023}
}

@article{bai2023qwen,
  title={Qwen-vl: A frontier large vision-language model with versatile abilities},
  author={Bai, Jinze and Bai, Shuai and Yang, Shusheng and Wang, Shijie and Tan, Sinan and others},
  journal={arXiv preprint arXiv:2308.12966},
  volume={1},
  number={2},
  pages={3},
  year={2023}
}

@article{grattafiori2024llama,
  title={The llama 3 herd of models},
  author={Grattafiori, Aaron and Dubey, Abhimanyu and Jauhri, Abhinav and Pandey, Abhinav and Kadian, Abhishek and Al-Dahle, Ahmad and Letman, Aiesha and Mathur, Akhil and Schelten, Alan and Vaughan, Alex and others},
  journal={arXiv preprint arXiv:2407.21783},
  year={2024}
}

@article{team2023gemini,
  title={Gemini: a family of highly capable multimodal models},
  author={Team, Gemini and Anil, Rohan and Borgeaud, Sebastian and Alayrac, Jean-Baptiste and Yu, Jiahui and Soricut, Radu and Schalkwyk, Johan and Dai, Andrew M and Hauth, Anja and Millican, Katie and others},
  journal={arXiv preprint arXiv:2312.11805},
  year={2023}
}

@inproceedings{caron2021emerging,
  title={Emerging properties in self-supervised vision transformers},
  author={Caron, Mathilde and Touvron, Hugo and Misra, Ishan and J{\'e}gou, Herv{\'e} and Mairal, Julien and Bojanowski, Piotr and Joulin, Armand},
  booktitle={ICCV},
  pages={9650--9660},
  year={2021}
}

@article{sethian1996fast,
  title={A fast marching level set method for monotonically advancing fronts.},
  author={Sethian, James A},
  journal={Proceedings of the National Academy of Sciences},
  volume={93},
  number={4},
  pages={1591--1595},
  year={1996}
}

@inproceedings{tolias2013aggregate,
  title={To aggregate or not to aggregate: Selective match kernels for image search},
  author={Tolias, Giorgos and Avrithis, Yannis and J{\'e}gou, Herv{\'e}},
  booktitle={ICCV},
  pages={1401--1408},
  year={2013}
}

@inproceedings{wang2025vggt,
  title={{VGGT}: {V}isual geometry grounded transformer},
  author={Wang, Jianyuan and Chen, Minghao and Karaev, Nikita and Vedaldi, Andrea and Rupprecht, Christian and Novotny, David},
  booktitle={CVPR},
  pages={5294--5306},
  year={2025}
}

@article{vggt-slam,
  title={{VGGT-SLAM: Dense RGB SLAM optimized on the SL(4) manifold}},
  author={Maggio, Dominic and Lim, Hyungtae and Carlone, Luca},
  journal={arXiv preprint arXiv:2505.12549},
  year={2025}
}

@inproceedings{murai2025mast3r-slam,
  title={{MASt3R-SLAM: Real-time dense SLAM with 3D reconstruction priors}},
  author={Murai, Riku and Dexheimer, Eric and Davison, Andrew J},
  booktitle={CVPR},
  pages={16695--16705},
  year={2025}
}

@inproceedings{wang2024dust3r,
  title={Dust3r: Geometric 3d vision made easy},
  author={Wang, Shuzhe and Leroy, Vincent and Cabon, Yohann and Chidlovskii, Boris and Revaud, Jerome},
  booktitle={CVPR},
  pages={20697--20709},
  year={2024}
}

@inproceedings{leroy2024grounding,
  title={Grounding image matching in 3d with mast3r},
  author={Leroy, Vincent and Cabon, Yohann and Revaud, J{\'e}r{\^o}me},
  booktitle={ECCV},
  pages={71--91},
  year={2024}
}

@inproceedings{yang2025fast3r,
  title={Fast3r: Towards 3d reconstruction of 1000+ images in one forward pass},
  author={Yang, Jianing and Sax, Alexander and Liang, Kevin J and Henaff, Mikael and Tang, Hao and Cao, Ang and Chai, Joyce and Meier, Franziska and Feiszli, Matt},
  booktitle={CVPR},
  pages={21924--21935},
  year={2025}
}

@article{brown2020language,
  title={Language models are few-shot learners},
  author={Brown, Tom and Mann, Benjamin and Ryder, Nick and Subbiah, Melanie and Kaplan, Jared D and Dhariwal, Prafulla and Neelakantan, Arvind and Shyam, Pranav and Sastry, Girish and Askell, Amanda and others},
  journal={NeurIPS},
  volume={33},
  pages={1877--1901},
  year={2020}
}

@inproceedings{zhou2025nolo,
  title={Nolo: Navigate only look once},
  author={Zhou, Bohan and Zhang, Zhongbin and Wang, Jiangxing and Lu, Zongqing},
  booktitle={IROS},
  pages={17162--17169},
  year={2025}
}

@article{yu2025c,
  title={C-NAV: Towards Self-Evolving Continual Object Navigation in Open World},
  author={Yu, Ming-Ming and Zhu, Fei and Liu, Wenzhuo and Yang, Yirong and Wang, Qunbo and Wu, Wenjun and Liu, Jing},
  journal={arXiv preprint arXiv:2510.20685},
  year={2025}
}

@inproceedings{chen2025recoverable,
  title={Recoverable compression: A multimodal vision token recovery mechanism guided by text information},
  author={Chen, Yi and Xu, Jian and Zhang, Xu-Yao and Liu, Wen-Zhuo and Liu, Yang-Yang and Liu, Cheng-Lin},
  booktitle={AAAI},
  pages={2293--2301},
  year={2025}
}

@article{yu2024vln,
  title={Vln-game: Vision-language equilibrium search for zero-shot semantic navigation},
  author={Yu, Bangguo and Liu, Yuzhen and Han, Lei and Kasaei, Hamidreza and Li, Tingguang and Cao, Ming},
  journal={arXiv preprint arXiv:2411.11609},
  year={2024}
}
\end{document}